\documentclass[12pt]{article}
\usepackage{amsmath,bbm,amssymb}
\usepackage{amsthm}
\usepackage{comment}
\usepackage{graphicx, graphics}
\usepackage{textcomp}
\usepackage{amsfonts}

\usepackage{psfrag,pst-node,subfigure,rotating, amsmath, bbm, amsthm, amssymb, amsthm, setspace, picture, epsfig, amsfonts, upgreek}

\usepackage{multirow}

\usepackage{arydshln}

\newtheorem{theorem}{Theorem}

\def\v{\mathbf}\def\m{\mathbf}\def\vg{\boldsymbol}\def\s{\mathcal}

\begin{document}
\begin{center}
    {\Large\bf The Reconstruction Approach: From Interpolation to Regression}
\\[2mm] {\large Shifeng Xiong}
\\ NCMIS, Academy of Mathematics and Systems Science \\Chinese Academy of Sciences, Beijing 100190\\xiong@amss.ac.cn
\end{center}

\vspace{1cm} \noindent{\bf Abstract}\quad This paper introduces an interpolation-based method, called the reconstruction approach, for nonparametric regression. Based on the fact that interpolation usually has negligible errors compared to statistical estimation, the reconstruction approach uses an interpolator to parameterize the regression function with its values at finite knots, and then estimates these values by (regularized) least squares. Some popular methods including kernel ridge regression can be viewed as its special cases. It is shown that, the reconstruction idea not only provides different angles to look into existing methods, but also produces new effective experimental design and estimation methods for nonparametric models. In particular,  for some methods of complexity $O(n^3)$, where $n$ is the sample size, this approach provides effective surrogates with much less computational burden. This point makes it very suitable for large datasets.


\vspace{1cm} \noindent{{\bf KEY WORDS:} Gaussian process regression; Kernel method; Kriging; Smoothing.}



\section{Introduction}\label{sec:intro}

Nonparametric regression is one of core issues in statistics and machine learning. There are two basic classes of methods for regression function estimation: local methods and parameterization methods. The first class includes local polynomial regression methods (Fan and Gijbels 1996), nearest-neighbor methods (Dasarathy 1991), and tree methods (Breiman et al. 1984). The second class represents the unknown function in an infinite-dimensional space as a form with finite unknown parameters, and then the problem is approximately transformed to a parametric one. The main manner of parameterization is basis representation that uses basis function expansions to replace unknown functions. Popular basis functions include polynomials, splines (Eubank 1999), and kernel bases (Berlinet and Thomas-Agnan 2004). Neural networks (Goodfellow, Bengio, and Courville 2016) that represent functions as complex composite parametric forms also belong to parameterization methods. It is known that the nonparametric regression problem is still valuable to investigate, especially in the dimensions much higher than two or three (Friedman, Hastie, and Tibshirani 2008).

Interpolation is an important technique for function approximation, and has been intensively studied by mathematicians (Wendland 2004). It can be viewed as the limit of a regression problem as noises go to zeros, and iterative regression techniques have been used to approximate an interpolator (Friedman 2001; Kang and Joseph 2016). Also, many techniques used in regression are applicable to interpolation such as basis representation. In statistics, interpolation is commonly used to model some spatial data (Cressie 2015), functional data (Ramsay, Hooker, and Graves 2009), and computer experiments (Santner, Williams, and Notz 2003), which do not contain any random noise. For noisy data, applications of interpolation are very limited. It is sometimes served as an auxiliary technique in nonparametric regression (Hall and Turlach 1997).

\begin{figure}[tbp]\begin{center}
\scalebox{0.6}[0.6]{\includegraphics{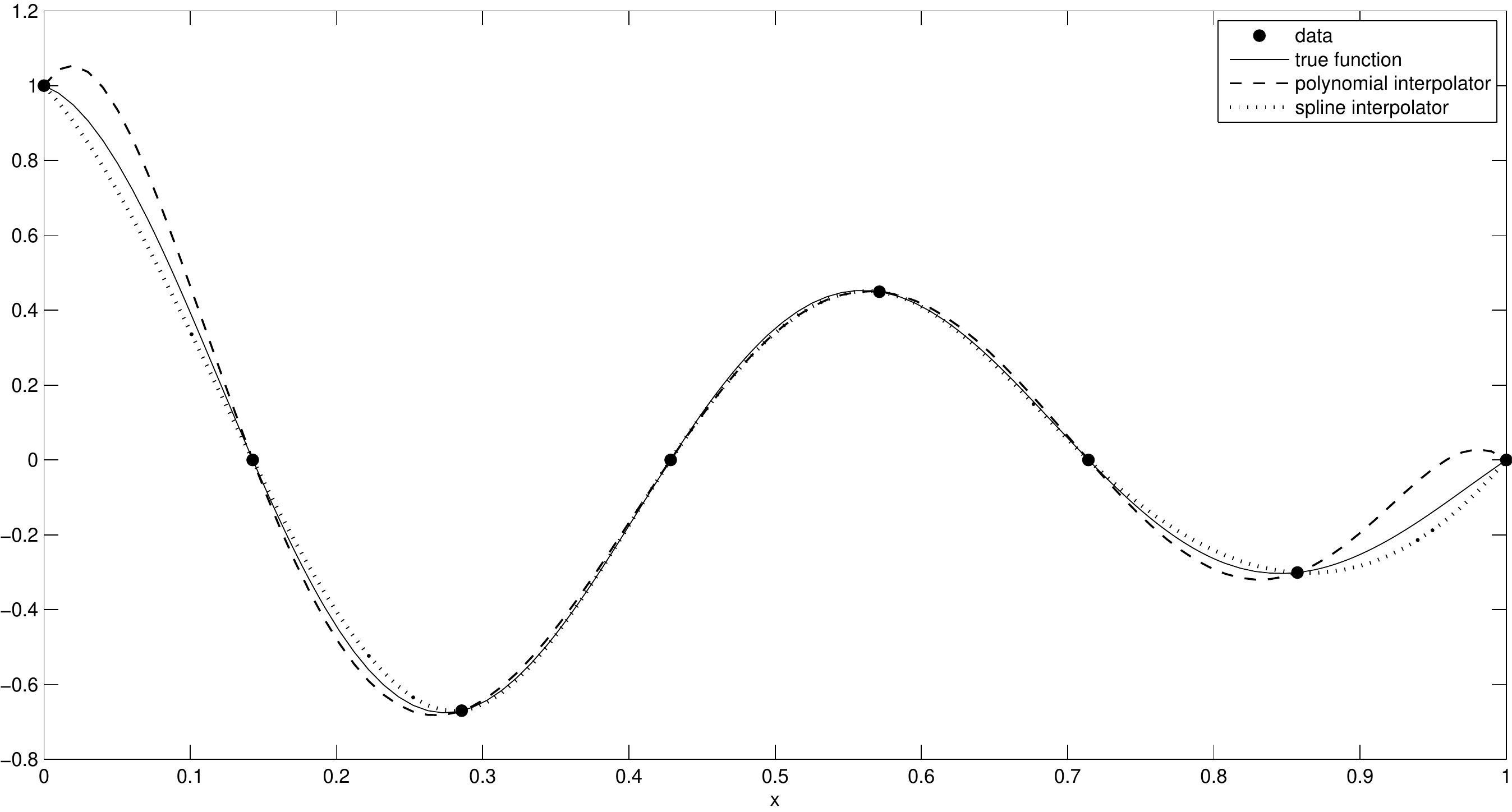}}
\end{center}
\caption{An example of interpolation.}\label{fig:intpt}
\end{figure}

Usually the convergence of an interpolator to the true function is quite fast. We take the following function, \begin{equation}\label{f1d}f(x)=\exp(-1.4x)\cos(3.5\pi x),\quad x\in[0,1],\end{equation} in Santner, Williams, and Notz (2003) for example. With only eight observations at equally spaced knots, polynomial interpolation and cubic spline interpolation both yield satisfactory approximation (Figure \ref{fig:intpt}), while a regression estimator based on noisy data usually requires much more observations to reach similar accuracy. In fact, popular interpolators can converge at high-order power rates, even at exponential rates, for sufficiently smooth functions (Stewart 1996; Wendland 2004) (Roughly, here a $d$-dimensional sufficiently smooth function is defined as a function with $\nu\gg d$, where $\nu$ denotes its degree of smoothness). Such rates are much faster than that of statistical estimation restricted by the central limit theorem. For estimating a regression function with noisy data, that fact motivates us to represent the unknown function as its interpolator at some knots. Such a representation is reasonable since the difference between the function itself and its interpolator is negligible compared to statistical errors. Consequently, the parameters that we need to estimate are the values of the unknown function at these knots, and can be estimated by minimizing a (regularized) quadratic risk function. We call this procedure the reconstruction approach since it can be viewed as a process of reconstructing the whole function with its finite values.

The reconstruction approach is a general method for multiple nonparametric regression. It is a parameterization method, and its main difference from existing parameterization methods is the clear interpretation of the parameters: they are the function values at the knots. This point facilitates the estimation process in some cases. Besides, in this paper we will show several appealing features of the proposed approach. First, it provides different angles to look into popular methods such as polynomial regression  (Celant and Broniatowski 2016) and kernel ridge regression  (Saunders, Gammerman, and Vovk 1998). These methods can be viewed as special cases of the reconstruction approach with according re-parameterization strategies. Second, the reconstruction approach focuses on estimating the function values at selected knots, and this idea can lead to new statistical methods. We will present new experimental design and estimation methods for nonparametric models, and will show their effectiveness with numerical examples. In particular, it allows selection of a small number of knots, and this point makes it suitable for large datasets. Third, the proposed approach builds a systematic connection between interpolation and regression. If any feasible interpolation method appears, then it can be used in the proposed approach to construct new regression estimation. The reconstruction approach broadens the applications of interpolation.

This paper is organized as follows. Section \ref{sec:nr} gives a general description of the reconstruction approach. Section \ref{sec:1d} discusses the reconstruction approach in one-dimensional regression models. Section \ref{sec:kra} studies Gaussian process reconstruction methods for multiple regression problems. Section \ref{sec:simu} presents numerical examples for Gaussian process reconstruction regression. We conclude the paper with some discussion in Section \ref{sec:dis}. Technical proofs are given in the the on-line Supplementary Materials.

\section{General description of the reconstruction approach}\label{sec:nr}

Consider the nonparametric regression model \begin{equation}y_i=f(\v{x}_i)+\varepsilon_i,\quad i=1,\ldots,n,\label{nr}\end{equation}where $f$ is a sufficiently smooth function defined on $[0,1]^d$, $\v{x}_i=(x_{i1},\ldots,x_{id})'\in[0,1]^d$ for $i=1,\ldots,n$, $\m{A}'$ denotes the transpose of a vector or matrix $\m{A}$, and $\varepsilon_i$'s are independent random errors with $E\varepsilon_i=0$ and $E\varepsilon_i^2=\sigma^2<\infty$ for $i=1,\ldots,n$.
We need to estimate the unknown regression function $f$ based on training data $\{(\v{x}_1,y_1),\ldots,(\v{x}_n,y_n)\}$. Write $\s{X}=\{\v{x}_1,\ldots,\v{x}_n\}$. For a set of knots $\s{A}=\{\v{a}_1,\ldots,\v{a}_m\}\subset[0,1]^d$ and $\vg{\gamma}=(\gamma_1,\ldots,\gamma_m)'\in{\mathbb{R}}^m$, let $\mathcal{I}(\v{x};\,\s{A},\vg{\gamma})$ be an interpolator on $[0,1]^d$, i.e., $\mathcal{I}(\v{a}_i;\,\s{A},\vg{\gamma})=\gamma_i$ for $i=1,\ldots,m$. Write $f_{\s{A}}=(f(\v{a}_1),\ldots,f(\v{a}_m))'$. Since $f$ can be approximately reconstructed as $\mathcal{I}(\v{x};\,\s{A},f_{\s{A}})$, we estimate $f$ by \begin{equation}\label{gf0}\hat{f}(\v{x})=\mathcal{I}(\v{x};\,\s{A},\hat{\vg{\gamma}}),\end{equation} where $\hat{\vg{\gamma}}=(\hat{\gamma},\ldots,\hat{\gamma}_m)'$ is the solution to \begin{equation}\label{gf}\min_{\vg{\gamma}\in{\mathbb{R}}^m}\frac{1}{n}\sum_{i=1}^n \big[y_i-\mathcal{I}(\v{x}_i;\,\s{A},\vg{\gamma})\big]^2+P(\vg{\gamma},\vg{\lambda}).\end{equation} Here $P$ is a penalty function, also called regularizer, with some tuning parameters $\vg{\lambda}$.

Equations \eqref{gf0} and \eqref{gf} give the general formulas of the reconstruction approach. It can be seen that this approach is actually a parameterization method that approximates a function in an infinity-dimensional space by a model of finite parameters. Furthermore, if the interpolator has a linear representation $\mathcal{I}(\v{x};\,\s{A},{\vg{\gamma}})={\vg{\gamma}}'\v{b}(\v{x})$, then our approach belongs to the class of basis representation-based nonparametric modeling methods. With popular basis functions, the reconstruction approach can be viewed as a re-parameterization of existing basis representation-based methods, and provides a different angle to look into them. Note that there is a clear interpretation of parameters to be estimated in the reconstruction approach: they are response values of $f$. From the reconstruction perspective, we can derive new experimental design and modeling methods that focus on the estimation of response values at selected knots. They will be discussed in the following sections.

The following theorem gives a simplified but intuitive interpretation of the theoretical validity of the reconstruction approach.
\begin{theorem}\label{th:rate}
Suppose that $\mathcal{I}(\v{x};\,\s{A},{\vg{\gamma}})={\vg{\gamma}}'\v{b}(\v{x})=\sum_{j=1}^m{\gamma}_jb_j(\v{x})$ with $\sup_{\v{x}\in[0,1]^d}\sum_{j=1}^m|b_j(\v{x})|<\infty$, and that $\hat{f}$ and $\hat{\vg{\gamma}}$ are given in \eqref{gf0} and \eqref{gf}, respectively. Let $\delta_m=\sup_{\v{x}\in[0,1]^d}|\mathcal{I}(\v{x};\,\s{A},f_{\s{A}})-f(\v{x})|$. If $E\sup_{j=1,\ldots,m}|\hat{{\gamma}}_j-f(\v{a}_j)|^2=O(\varepsilon_n)$, then for $\v{x}_0\in[0,1]$, the mean squared error (MSE) of $\hat{f}(\v{x}_0)$ is $$\mathrm{MSE}\left(\hat{f}(\v{x}_0)\right)=O\left(\delta_m^2+\varepsilon_n\right).$$\end{theorem}

Note that $\delta_m$ represents the convergence rate of the interpolator $\mathcal{I}$ to $f$. By Theorem \ref{th:rate}, if $\delta_m^2=O(\varepsilon_n)$, then the systematic error caused by the interpolator is negligible, and $\mathrm{MSE}\left(\hat{f}(\v{x}_0)\right)=O(\varepsilon_n)$, the same as the rate of $\hat{f}$ on $\s{A}$. In other words, for estimating the whole $f$, it suffices to consider the estimation of $f_{\s{A}}$. This is the core idea of the reconstruction approach. Furthermore, many interpolators have much faster convergence rates than $\varepsilon_n$, which is restricted by the optimal statistical rate $1/n$. We can therefore use a small $m$, even $m\ll n$, to satisfy $\delta_m^2=O(\varepsilon_n)$. Note that smaller $m$ generally means less computational cost. This point is important when the computation is a problem.

To illustrate the proposed approach, consider a linear interpolator $\mathcal{I}(\v{x};\,\s{A},{\vg{\gamma}})={\vg{\gamma}}'\v{b}(\v{x})$. Taking the quadratic penalty $P(\vg{\gamma},{\lambda})=\lambda\vg{\gamma}'\m{\Sigma}\vg{\gamma}$ with a semi-positive definite matrix $\m{\Sigma}$ in \eqref{gf}, we have
\begin{equation}\min_{\vg{\gamma}\in{\mathbb{R}}^n}\frac{1}{n}\sum_{i=1}^n\big[y_i-\vg{\gamma}'\v{b}(\v{x}_i)\big]^2+\lambda\vg{\gamma}'\m{\Sigma}\vg{\gamma},\label{rre0}
\end{equation}which implies
\begin{equation}
\hat{\vg{\gamma}}=(\m{B}'\m{B}+n\lambda\m{\Sigma})^{-1}\m{B}'\v{y}\ \ \text{and}\ \ \hat{f}(\v{x})=\v{y}'\m{B}(\m{B}'\m{B}+n\lambda\m{\Sigma})^{-1}\v{b}(\v{x}),\label{rre}\end{equation} where $\m{B}=(\v{b}(\v{x}_1),\ldots,\v{b}(\v{x}_n))'\in{\mathbb{R}}^{n\times m}$ and $\v{y}=(y_1,\ldots,y_n)'$. The tuning parameter $\lambda$ can be selected by minimizing the generalized cross-validation (GCV) criterion (Golub, Heath, and Wahba 1979),
\begin{equation}\label{gcv0}{\mathrm{GCV}}(\lambda)=\frac{\|\v{y}-\m{B}(\m{B}'\m{B}+n\lambda\m{\Sigma})^{-1}\m{B}'\v{y}\|^2}{n\big[1-{\mathrm{trace}}(\m{B}(\m{B}'\m{B}
+n\lambda\m{\Sigma})^{-1}\m{B}')/n\big]^2},
\end{equation}or other CV-like criteria.

\section{Reconstruction in one-dimensional regression}\label{sec:1d}

\subsection{Replication design}\label{subsec:rd}

Consider the one-dimensional version of \eqref{nr}, \begin{equation}y_i=f(x_i)+\varepsilon_i,\quad i=1,\ldots,n,\label{snr}\end{equation}where $f$ is a sufficiently smooth function defined on $[0,1]$.
In many applications, these $x_1,\ldots,x_n$ can be designed to maximize the accuracy of the fitted function (Box and Draper 2007). A popular design under the nonparametric model is to assign $x_i$'s to be equally spaced, i.e., $x_i=(i-1)/(n-1)$ for $i=1,\ldots,n$. Such a design is also called uniform design and is optimal with respect to some criterion (Xie and Fang 2000).

As shown in the previous section, we can use a small $m$ from the reconstruction idea. We now consider the following replication design of $x_1,\ldots,x_n$ on $\s{A}=\{a_1,\ldots,a_m\}\subset[0,1]$ with a small $m$. Assign $l$ replications at each knot, i.e., $x_1=\cdots=x_l=a_1,\ldots,x_{(m-1)l+1}=\cdots=x_n=a_m$, and then reconstruct $f$ by \eqref{gf0} with an interpolator $\mathcal{I}({x};\,\s{A},f_{\s{A}})$. With the quadratic loss and without a penalty in \eqref{gf}, we have that $\hat{\vg{\gamma}}$ in \eqref{gf0} has a simple form $\left((y_1+\cdots+y_l)/l,\ldots,(y_{(m-1)l+1}+\cdots+y_n)/l\right)'$.

When $\mathcal{I}$ is selected as the polynomial interpolator (De Boor 1978), the estimator has the Lagrange form \begin{equation}\label{pi}\hat{f}(x)=\sum_{j=1}^m\left[\hat{\gamma}_j\left(\prod_{1\leqslant k\leqslant m,\ k\neq j}\frac{x-a_k}{a_j-a_k}\right)\right].\end{equation}It can be seen that this form is equivalent to that of traditional polynomial regression $$\hat{f}(x)=\hat{\beta}_1+\hat{\beta}_2x+\cdots+\hat{\beta}_{m}x^{m-1},$$where the $\hat{\beta}_j$'s are estimated by the least squares (without a penalty) and can be represented as linear forms of $\hat{\gamma}_j$'s. Therefore, here the reconstruction approach does not produce a new regression estimator. However, from the reconstruction angle, we can derive the convergence rate of $\hat{f}$ in \eqref{pi} based on the interpolation theory like Theorem \ref{th:rate}, while there is limited similar results on polynomial regression in the literature (Eubank 1999). Here we use the Chebyshev nodes \begin{equation}\label{cn}\s{A}=\{a_j=1/2-\cos[(2j-1)\pi/2m]/2:\,j=1,\ldots,m\}\end{equation} to avoid Runge's Phenomenon (De Boor 1978). Under some conditions, for $x_0\in[0,1]$,
\begin{equation}\label{pir}\mathrm{MSE}\left(\hat{f}(x_0)\right)=O\left({\frac{\log(n)}{n}}\right)\quad \text{or}\quad o\left({\frac{\log(n)}{n}}\right),\end{equation}which is very close to a parametric rate of convergence. The proof of \eqref{pir} can be found in the Supplementary Materials. It should be pointed out that there are systematic results on the error analysis and optimal design of the polynomial model, i.e., $f$ in \eqref{snr} is indeed a polynomial of order $m-1$ (Celant and Broniatowski 2016). The Chebyshev nodes in \eqref{cn} are also useful for that parametric setting.

We next consider the cubit spline interpolator (De Boor 1978). It is also a linear interpolator, and its computation can be found in many textbooks such as De Boor (1978). Let \begin{equation}\label{erd}\s{A}=\{a_j=(j-1)/(m-1):\,j=1,\ldots,m\}.\end{equation} Similar to \eqref{pir}, we can obtain that, under some conditions,
\begin{equation}\label{sir}\mathrm{MSE}\left(\hat{f}(x_0)\right)=O\left(n^{-8/9}\right),\quad x_0\in[0,1],\end{equation} and the proof is deferred in the Supplementary Materials. This rate is consistent with that of popular local regression methods with appropriately selected kernel functions (Eubank 1999).

\begin{figure}[tbp]\begin{center}
\scalebox{0.6}[0.6]{\includegraphics{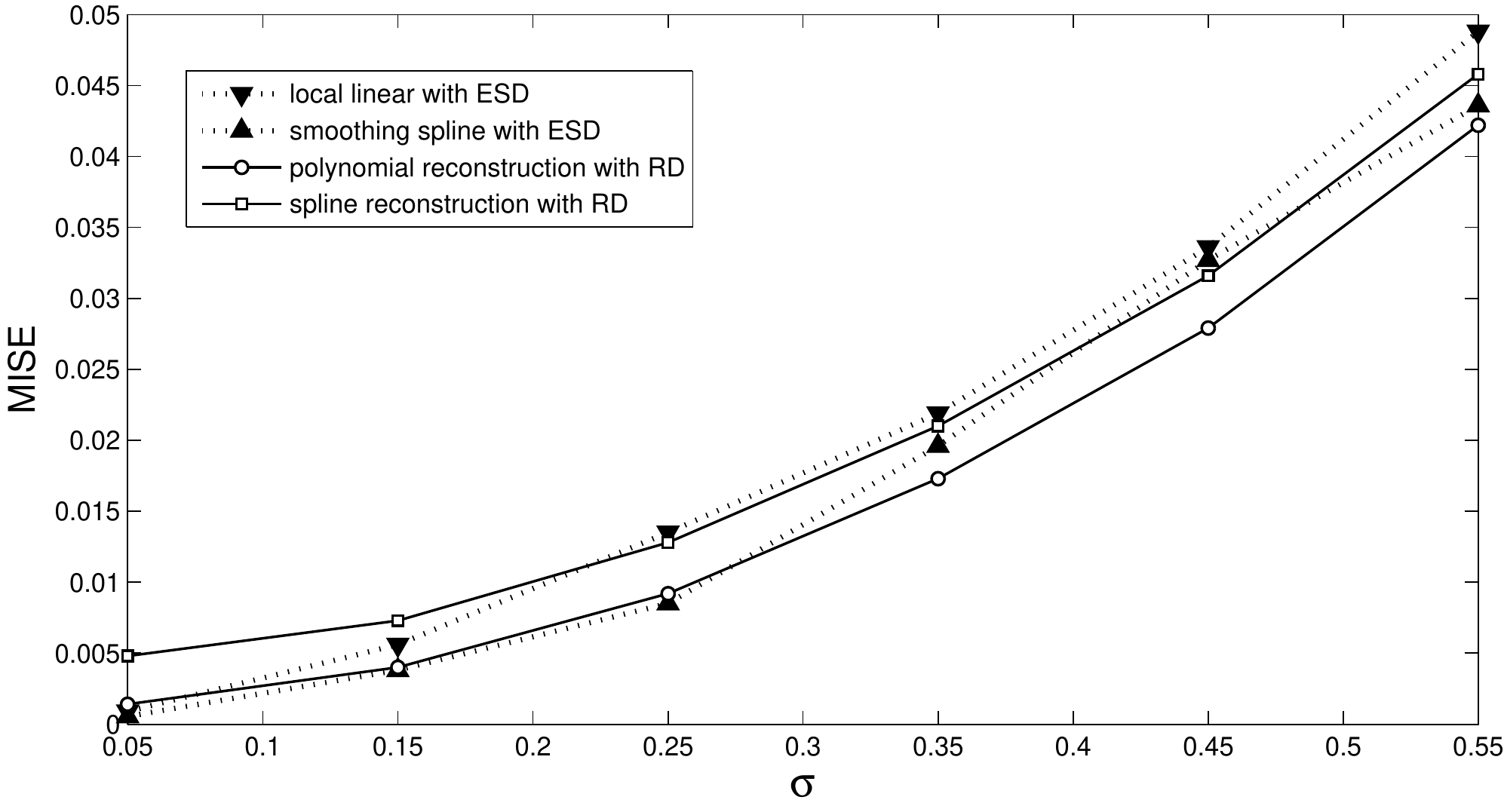}}
\end{center}
\caption{MISE comparisons for estimators with the replication designs (RDs) and equally spaced designs (ESDs).}\label{fig:cd}
\end{figure}

A simulation is conducted to compare the proposed replication designs and traditional equally spaced designs. With $n=49$, we compute the local linear estimator and smoothing spline estimator based on the equally spaced design $x_i=(i-1)/(n-1),\ i=1,\ldots,n$, the polynomial interpolation-based reconstruction estimator based on the replication design \eqref{cn} with $m=l=7$, and the cubic spline interpolation-based reconstruction estimator based on the replication design \eqref{erd} with $m=l=7$. GCV is used to select the smoothing parameters in the local linear estimator and smoothing spline estimator. The regression function $f$ is set as in \eqref{f1d}. The standard deviation $\sigma$ of the random error varies from $0.05$ to $0.55$. The mean integrated squared errors (MISEs) over 100 replications are reported in Figure \ref{fig:cd}. Basically, the polynomial reconstruction estimator and spline reconstruction estimator with the corresponding replication designs are comparable to the smoothing spline estimator and local linear estimator with the equally spaced design, respectively. When $\sigma$ is small, the MISEs of the replication design-based estimators are slightly larger than the equally spaced design-based estimators. The reason is that, compared with the statistical estimation errors that are proportion to the small $\sigma$, the relatively small number of knots leads to relatively large biases from the interpolation techniques. As $\sigma$ increases, the replication design-based estimators become relatively effective since the biases from interpolation tend to be negligible. Besides the comparable efficiency, the proposed replication design-based estimators possess some advantages. Replication designs can detect heteroscedasticity straightforwardly. For some practical cases, experiments at less sites ($m\ll n$) can save much cost.

\subsection{Finite difference penalization}\label{subsec:fdp}

This subsection presents a new regression estimator based on the equally spaced design. Let $\s{A}=\s{X}$. In the reconstruction approach, we need to estimate $n$ parameters $\vg{\gamma}=f_{\s{X}}$, and then combine their estimators with an interpolator. Recall that the (cubic) smoothing spline method uses a penalty $\lambda\int_0^1 [f''(x)]^2dx$ to control the roughness of $f$ (Gu 2002). Here a sum of finite differences is used to approximate such a penalty, and we estimate $\vg{\gamma}$ by \begin{equation}\label{fdp}\min_{\vg{\gamma}\in{\mathbb{R}}^n}\frac{1}{n}\|\v{y}-\vg{\gamma}\|^2+\lambda\sum_{i=2}^{n-1}(\gamma_{i+1}-2\gamma_i+\gamma_{i-1})^2,\end{equation}where $\lambda>0$ is a tuning parameter.
The solution to this problem has a closed form \begin{equation}\label{fdpe}\hat{\vg{\gamma}}=(n\lambda\m{M}'\m{M}+\m{I}_n)^{-1}\v{y},\end{equation}where $$\m{M}=\left(\begin{array}{cccccccc}1&-2&1&0&\cdots&0&0&0\\0&1&-2&1&\cdots&0&0&0\\&&&&\ddots&&&\\0&0&0&0&\cdots&1&-2&1\end{array}\right)_{(n-2)\times n}.$$The tuning parameter $\lambda$ can be selected by miniziming
$${\mathrm{GCV}}(\lambda)=\frac{\|\v{y}-(n\lambda\m{M}'\m{M}+\m{I}_n)^{-1}\v{y}\|^2}{n\big[1-{\mathrm{trace}}((n\lambda\m{M}'\m{M}+\m{I}_n)^{-1})/n\big]^2}.$$

\begin{figure}[tbp]\begin{center}
\scalebox{0.6}[0.6]{\includegraphics{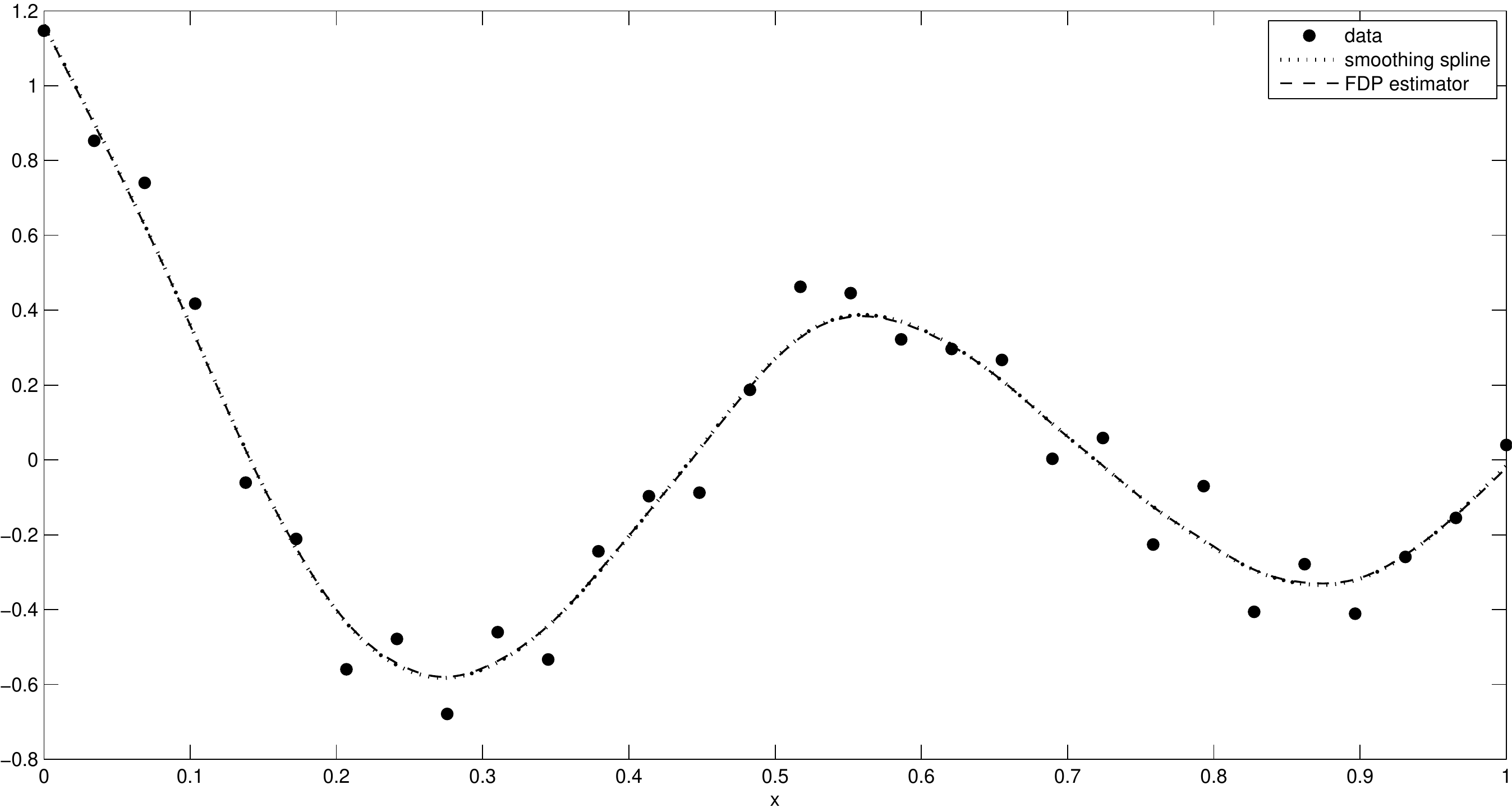}}
\end{center}
\caption{Comparison of the finite difference penalization (FDP) estimator and the smoothing spline estimator with 30 observations.}\label{fig:fdp}
\end{figure}

Since popular interpolation techniques with $n$ knots have negligible errors compared to statistical estimation (i.e. $\delta_n\ll \varepsilon_n$ in Theorem \ref{th:rate}), anyone of them can be used to reconstruct the whole estimator of $f$ based on $\hat{\vg{\gamma}}$ in \eqref{fdpe}.
Such a finite difference penalization estimator can be viewed as a discrete approximation to the smoothing spline method. Figure \ref{fig:fdp} compares the two estimators based on 30 observations that are randomly generated with $f$ in \eqref{f1d}. The cubic spline interpolator is used to reconstruct the whole estimator in the finite difference penalization method. The tuning parameters in the two methods are both selected by GCV. It can be seen that the two estimators almost coincide with each other.
Note that the computations in \eqref{fdpe} and in cubic spline interpolation only need to solve sparse linear systems (Stewart 1998), which have total complexity of $O(n)$ (Toraichi et al. 1987). They are much cheaper than to compute the inverse of a dense matrix in the smoothing spline method whose complexity is $O(n^3)$ (Gu 2002). Therefore, the finite difference penalization approach may act as a good substitute of the smoothing spline method when $n$ is very large.
It should be pointed out that the finite difference penalization estimator is also related to the trend filtering estimator (Tibshirani 2014) that uses $\ell_1$ regularizer in \eqref{fdp}. Another related method is the fused ridge estimator that is used to estimate the coefficients of a linear model (van Wieringen 2018).

\section{Gaussian process reconstruction regression}\label{sec:kra}

\subsection{Kernel interpolation and the Gaussian process model}\label{subsec:kc}

Kernel methods are commonly used in interpolation (Wendland 2004), statistics (Berlinet and Thomas-Agnan 2004), and machine learning (Kung 2014). This subsection provides a brief introduction of kernel interpolation and the related Gaussian process model.

Let $K:\ [0,1]^d\times[0,1]^d\mapsto \mathbb{R}$ be a symmetric positive definite kernel. Here we consider stationary kernels, i.e., $K(\v{x}_1,\v{x}_2)=R(\v{x}_1-\v{x}_2)$. Define the linear space $$F_R=\left\{\sum_{i=1}^n\beta_iR(\cdot-\v{x}_i):\ \beta_i\in\mathbb{R},\ \v{x}_i\in[0,1]^d,\ i=1,\ldots,n,\ \text{for all}\ n=1,2,\ldots\right\},$$and equip this space with the bilinear form
$$\left<\sum_{i=1}^n\beta_iR(\cdot-\v{x}_i),\ \sum_{j=1}^m\tilde{\beta}_iR(\cdot-\tilde{\v{x}}_j)\right>_R=\sum_{i=1}^n\sum_{j=1}^m\beta_i\tilde{\beta}_jR(\v{x}_i-\tilde{\v{x}}_j).$$The closure of $F_R$ under the inner product $\big<\cdot,\cdot\big>_R$ is a reproducing kernel Hilbert space, denoted by $\s{N}_R$, and the norm of $\s{N}_R$ is $\|f\|_{\s{N}_R}=\sqrt{\big<f,f\big>_{\s{N}_R}}$, where $f\in\s{N}_R$ and $\big<f,f\big>_{\s{N}_R}$ is induced by $\big<\cdot,\cdot\big>_R$.

Popular choices of $R$ include the Gaussian kernel \begin{equation}\label{gk}R(\v{h})=\exp\left(-\sum_{j=1}^d\theta_jh_{j}^2\right),\ \ \v{h}=(h_1,\ldots,h_d)',\end{equation} for fixed $\theta_j>0$ and the Mat\'{e}rn kernel
\begin{equation}\label{mk}R(\v{h})=\prod_{j=1}^d\frac{1}{\Gamma(\nu)2^{\nu-1}}\left(\frac{2\sqrt{\nu}|h_j|}{\phi}\right)^\nu K_\nu\left(\frac{2\sqrt{\nu}|h_j|}{\phi}\right),\ \ \v{h}=(h_1,\ldots,h_d)',\end{equation} for fixed $\nu>0$ and $\phi>0$, where $K_\nu$ is the modified Bessel function of order $\nu$. The reproducing kernel Hilbert space $\s{N}_R$ possesses the corresponding smoothness properties to the kernel $R$, and it relates to the Sobolev space (Wendland 2004). The space of polynomial functions and the space of spline functions are special cases of reproducing kernel Hilbert space with certain kernels.

The kernel interpolator (also called reproducing kernel Hilbert space interpolator) with respect to kernel $R$ is the solution to the optimization problem
\begin{eqnarray*}&&\min_{g\in\s{N}_R}\|g\|_{\s{N}_R},\\&&\text{s.t.}\ g(a_j)=\gamma_j\ j=1,\ldots,m,\end{eqnarray*}and has the closed form
\begin{equation}\label{ki}\mathcal{I}(\v{x};\,\s{A},\vg{\gamma})=\vg{\gamma}'\m{R}_{\s{A}}^{-1}\v{r}_{\s{A}}(\v{x}),\end{equation} where ${\v{r}}_{\s{A}}(\v{x})=\big(R(\v{x}-\v{a}_1),\ldots,R(\v{x}-\v{a}_m)\big)'$, and ${\m{R}}_{\s{A}}=\big(R(\v{a}_i-\v{a}_j)\big)_{i,j=1,\ldots,m}$. The convergence rate of the kernel interpolator \eqref{ki} to $f$ is well-established in the literature (Wendland 2004). For sufficiently smooth functions, it can converge at an exponential rate.

The kernel interpolator can also be derived under the Gaussian process model, also called the Kriging model (Matheron 1963), which is widely used in spatial statistics (Cressie 2015), computer experiments (Santner, Williams and Notz 2003), and machine learning (Rasmussen and Williams 2006). Here the kernel function $R$ is served as the correlation function of a Gaussian process. Specifically, assume that the unknown function $f$ follows
\begin{equation}f(\v{x})=\v{g}(\v{x})'\vg{\beta}+Z(\v{x}),\label{kriging}\end{equation}where $\v{g}(\v{x})=\left(g_1(\v{x}),\ldots,g_q(\v{x})\right)'$ is
a pre-specified set of regression functions, $\vg{\beta}$ is a vector of unknown regression coefficients, and $Z(\v{x})$ is a stationary
Gaussian process with mean zero, variance $\tau^2$, and correlation function $R$, denoted by GP$(0, \tau^2R)$. With the observations $\vg{\gamma}=(\gamma_1,\ldots,\gamma_m)'$ on $\s{A}=\{\v{a}_1,\ldots,\v{a}_m\}$, the best linear unbiased predictor (Santner, Williams and Notz 2003) is
\begin{eqnarray}\mathcal{I}_{\mathrm{GP}}(\v{x};\,\s{A},\vg{\gamma})
=\v{g}(\v{x})'\hat{\vg{\beta}}+{\v{r}_{\s{A}}(\v{x})}'{\m{R}}_{\s{A}}^{-1}\big(\vg{\gamma}-\m{G}_{\s{A}}\hat{\vg{\beta}}\big),\label{fity0}\end{eqnarray}
where ${\v{r}}_{\s{A}}$ and ${\m{R}}_{\s{A}}$ are the same as in \eqref{ki}, $\hat{\vg{\beta}}=(\m{G}_{\s{A}}'\m{R}_{\s{A}}^{-1}\m{G}_{\s{A}})^{-1}\m{G}_{\s{A}}'\m{R}_{\s{A}}^{-1}\vg{\gamma}$, and $\m{G}_{\s{A}}=\left(\v{g}(\v{a}_1),\ldots,\v{g}(\v{a}_m)\right)'$. We rewrite \eqref{fity0} as
\begin{equation}\mathcal{I}_{\mathrm{GP}}(\v{x};\,\s{A},\vg{\gamma})=\vg{\gamma}'\v{b}(\v{x}),\label{fity}\end{equation}where
$\v{b}(\v{x})=\m{U}\v{g}(\v{x})+\m{V}{\v{r}}_{\s{A}}(\v{x})$, $\m{U}=\m{R}_{\s{A}}^{-1}\m{G}_{\s{A}}(\m{G}_{\s{A}}'\m{R}_{\s{A}}^{-1}\m{G}_{\s{A}})^{-1}$, and \\$\m{V}=\left[\m{I}_m-\m{R}_{\s{A}}^{-1}\m{G}_{\s{A}}(\m{G}_{\s{A}}'\m{R}_{\s{A}}^{-1}\m{G}_{\s{A}})^{-1}\m{G}_{\s{A}}'\right]\m{R}_{\s{A}}^{-1}$.
It can be seen that $\mathcal{I}_{\mathrm{GP}}$ in \eqref{fity} is a linear interpolator with $m$ basis functions, which are linear combinations of $q+m$ basis functions $\v{g}(\v{x})$ and $\v{r}_{\s{A}}(\v{x})$. It reduces to the kernel interpolator \eqref{ki} when there is no regression function $\v{g}$ in \eqref{kriging}.

\subsection{Reconstruction regression via Gaussian process interpolation}\label{subsec:kcrc}

We use the Gaussian process interpolator \eqref{fity} in the reconstruction approach, and call it the Gaussian process reconstruction regression (GPRR) method. In this method, it is sometimes necessary to select a penalty that controls the roughness of the estimator for preventing overfitting, especially for large $m$ (Eilers and Marx 1996).
Note that $\|g\|_{\s{N}_R}$ describes the smoothness of $g\in\s{N}_R$ and recall that the parameterization of $f$ via \eqref{fity} is a sum of two parts, which are described by the basis functions $\v{g}(\v{x})$ and ${\v{r}}_{\s{A}}(\v{x})$, respectively. Usually $\v{g}(\v{x})$ is a vector of low-order polynomials. A natural penalty is the squared norm of the part corresponding to ${\v{r}}_{\s{A}}(\v{x})$,
\begin{equation}\label{kp}P(\vg{\gamma},\lambda)=\lambda\|\vg{\gamma}'\m{V}{\v{r}}_{\s{A}}(\v{x})\|_{\s{N}_R}^2=\lambda\vg{\gamma}'\m{V}\m{R}_{\s{A}}\m{V}'\vg{\gamma}.\end{equation} It therefore follows from \eqref{rre} that \begin{equation}\label{krrf}\hat{\vg{\gamma}}=(\m{B}'\m{B}+n\lambda\m{V}\m{R}_{\s{A}}\m{V}')^{-1}\m{B}'\v{y}\ \ \text{and}\ \ \hat{f}(\v{x})=\v{y}'\m{B}(\m{B}'\m{B}+n\lambda\m{V}\m{R}_{\s{A}}\m{V}')^{-1}\v{b}(\v{x}),\end{equation}where
\begin{equation}\label{bigB}\m{B}=(\v{b}(\v{x}_1),\ldots,\v{b}(\v{x}_n))'=\m{G}_{\s{X}}\m{U}'+\m{R}_{\s{X}\s{A}}\m{V},\end{equation} $\v{b}(\v{x})$, $\m{U}$, and $\m{V}$ are defined in \eqref{fity}, and ${\m{R}}_{\s{X}\s{A}}=\big(R(\v{x}_i-\v{a}_j)\big)_{i=1,\ldots,n,j=1,\ldots,m}$. The tuning parameter $\lambda$ can be selected by GCV \eqref{gcv0}.

Next we consider the important case of $\s{A}=\s{X}$. With the penalty \eqref{kp}, the optimization problem \eqref{rre0} reduces to \begin{equation*}\label{}\min_{\vg{\gamma}\in{\mathbb{R}}^n}\frac{1}{n}\sum_{i=1}^n[y_i-\vg{\gamma}'\v{b}(\v{x}_i)]^2+\lambda\vg{\gamma}'\m{V}\m{R}_{\s{X}}\m{V}'\vg{\gamma}
=\frac{1}{n}\|\v{y}-\vg{\gamma}\|^2+\lambda\vg{\gamma}'\m{V}\m{R}_{\s{X}}\m{V}'\vg{\gamma},\end{equation*}which yields\begin{equation}\label{gprr}\hat{\vg{\gamma}}=(\m{I}_n+n\lambda\m{V}\m{R}_{\s{X}}\m{V}')^{-1}\v{y}\ \ \text{and}\ \ \hat{f}(\v{x})=\v{y}'(\m{I}_n+n\lambda\m{V}\m{R}_{\s{X}}\m{V}')^{-1}\v{b}(\v{x}),\end{equation}
When using the special case \eqref{ki} of \eqref{fity}, we have the corresponding optimization problem \begin{equation}\label{krr0}\min_{\vg{\gamma}\in{\mathbb{R}}^n}\frac{1}{n}\sum_{i=1}^n[y_i-\vg{\gamma}'\m{R}_{\s{X}}^{-1}\v{r}_{\s{X}}(\v{x}_i)]^2+\lambda\vg{\gamma}'\m{R}_{\s{X}}^{-1}\vg{\gamma}
=\frac{1}{n}\|\v{y}-\vg{\gamma}\|^2+\lambda\vg{\gamma}'\m{R}_{\s{X}}^{-1}\vg{\gamma},\end{equation}which yields the kernel ridge regression (KRR) estimator (Saunders, Gammerman, and Vovk 1998) \begin{equation}\label{krr}\hat{f}(\v{x})=\v{y}'(\m{R}_{\s{X}}+n\lambda\m{I}_n)^{-1}\v{r}_{\s{X}}(\v{x}).\end{equation}
It can be seen that \eqref{krr0} is a re-parameterization of usual KRR formula, and can be used to understand KRR from the reconstruction viewpoint.

The KRR method can be derived from the Gaussian process regression (GPR) model, which adds a measurement error term to the interpolation model \eqref{kriging}. It is known that we need to compute the inverse of the $n\times n$ covariance matrix in GPR and KRR, and the complexity is of $O(n^3)$. Theorem \ref{th:rate} tells us that it suffices to use $m\ll n$ knots in our GPRR approach, which reduces the complexity to $O(m^2n)$. With such $m$ there is no need to use a penalty, and this further simplifies the computation. Therefore, GPRR with $m\ll n$ knots can be acted as a surrogate of GPR for large $n$. Note that many authors have proposed methods to speed up GPR for such cases, including the Nystr\"{o}m method and its modifications (Williams and Seeger 2000; Cressie and Johannesson 2008), tapering (Furrer, Genton, and Nychka 2006), the sparse pseudo-input Gaussian process (SPGP) method (Snelson and Ghahramani 2006), full scale approximation (Sang and Huang 2012), and local approximation (Gramacy and Apley 2015). Among them, the Nystr\"{o}m method and SPGP also have the complexity of $O(m^2n)$ with a similar meaning $m$. We will compare the two methods and GPRR via simulations in Section \ref{sec:simu}.

\subsection{A Bayesian viewpoint and the SPGP method}\label{subsec:bi}

The GPRR method has an empirical Bayesian interpretation as follows.
Assume that $f$ has the prior $f\sim$ GP$(0, \tau^2R)$, which is independent of the random errors $\varepsilon_i\sim N(0,\sigma^2),\ i=1,\ldots,n$.
Recall that $\vg{\gamma}=f_{\s{A}}=(f(\v{a}_1),\ldots,f(\v{a}_m))'$. For given $\v{x}$, an unbiased estimator of $f(\v{x})$ is
\begin{equation}E(f(\v{x})\,|\,\vg{\gamma})=\vg{\gamma}'\m{R}_{\s{A}}^{-1}\v{r}_{\s{A}}(\v{x}).\label{be}\end{equation}
Assume that $\tau^2$ and $\sigma^2$ are known. To estimate $\vg{\gamma}$ in \eqref{be}, we note that \begin{equation*}
\begin{pmatrix}\v{y}\\ \vg{\gamma}\end{pmatrix} \sim  N\left[\v{0},
\begin{pmatrix}
\tau^{2}\m{R}_{\s{X}}+\sigma^2\m{I}_n & \tau^{2}\m{R}_{\s{X}\s{A}}\\
\tau^{2}\m{R}_{\s{X}\s{A}}'& \tau^{2}\m{R}_{\s{A}}
 \end{pmatrix}\right],
\end{equation*}
which implies $$\v{y}\,|\,\vg{\gamma}\sim N\big[\m{R}_{\s{X}\s{A}}\m{R}_{\s{A}}^{-1}\vg{\gamma},\,
\tau^2(\m{R}_{\s{X}}-\m{R}_{\s{X}\s{A}}\m{R}_{\s{A}}^{-1}\m{R}_{\s{X}\s{A}}')+\sigma^2\m{I}_n\big].$$We use the following quasi-posterior mode to estimate $\vg{\gamma}$,
\begin{equation*}\max_{\vg{\gamma}\in{\mathbb{R}}^m}\ [\v{y}\,|\,\vg{\gamma}]_{\mathrm{q}}\times[\vg{\gamma}],\end{equation*}where $[\v{y}\,|\,\vg{\gamma}]_{\mathrm{q}}$ denotes the density of $N(\m{R}_{\s{X}\s{A}}\m{R}_{\s{A}}^{-1}\vg{\gamma},\,\sigma^2\m{I}_n)$, which replaces the covariance matrix in $[\v{y}\,|\,\vg{\gamma}]$ with $\sigma^2\m{I}_n$. This problem is equivalent to $$\min_{\vg{\gamma}\in{\mathbb{R}}^m}\|\v{y}-\m{R}_{\s{X}\s{A}}\m{R}_{\s{A}}^{-1}\vg{\gamma}\|^2+\sigma^2\vg{\gamma}'\m{R}_{\s{A}}^{-1}\vg{\gamma}/\tau^2.$$Plugging the estimator of $\vg{\gamma}$ obtained by the above equation into \eqref{be}, we get an empirical Bayesian estimator of $f(\v{x})$, which is actually the GPRR estimator in the previous subsection with the interpolator \eqref{ki} and the natural penalty \eqref{kp} for $\lambda=\sigma^2/(n\tau^2)$. From this viewpoint, we find that the idea of the SPGP method (Snelson and Ghahramani 2006) is very similar: it replaces the covariance matrix in $[\v{y}\,|\,\vg{\gamma}]$ with $\tau^2\vg{\Lambda}+\sigma^2\m{I}_n$, where $\vg{\Lambda}$ is the diagonal matrix of $\m{R}_{\s{X}}-\m{R}_{\s{X}\s{A}}\m{R}_{\s{A}}^{-1}\m{R}_{\s{X}\s{A}}'$.

It is known that the best estimator of $f(\v{x})$ in terms of MSE is $E(f(\v{x})\,|\,\v{y})=\v{y}'\big[\m{R}_{\s{X}}+\sigma^2\m{I}_n/(n\tau^2)\big]^{-1}\v{r}_{\s{X}}(\v{x})$, which corresponds to the KRR estimator with $\lambda=\sigma^2/(n\tau^2)$ in \eqref{krr}.
Only when $\s{A}=\s{X}$, the above empirical Bayesian estimator, or the SPGP estimator, is the best estimator $E(f(\v{x})|\v{y})$. With $\s{A}$ of $m\ll n$, an advantage of the empirical Bayesian and SPGP estimators is that they only need to compute inverses of $n\times n$ diagonal matrices.

\subsection{Selection of $\s{A}$}\label{subsec:alg}

A key issue in the reconstruction approach is the specification of the knot set $\s{A}$. For small or moderate $n$, we can select $\s{A}=\s{X}$. For large $n$, we hope to select a relatively small $m$ which still leads to satisfactory estimation accuracy since larger $m$ generally corresponds to heavier computations. A common rule in Gaussian process interpolation is to use the sample size of $10d$ (Loeppky, Sacks, and Welch 2009). We therefore recommend using \begin{equation}\label{m}m=10d.\end{equation} Since there is no strong guarantee that such a selection in GPRR is satisfactory, we can use some sequential strategies that add knots to re-estimate $f$, and \eqref{m} can be served as an initial point of $m$.

We now discuss the construction of $\s{A}$ for given $m$. First, we recommend using a subset of $\s{X}$ as $\s{A}$.  The main reason is that there are responses at such $\s{A}$, and thus the estimators of $\vg{\gamma}$ can use these responses as good starts in iterative algorithms for estimating $\vg{\gamma}$. In addition, the popular method KRR is a special case of the reconstruction approach with $\s{A}=\s{X}$; it can also be viewed as the limit of the reconstruction approach with $\s{A}\subset\s{X}$ as $m$ tends to $n$.
Second, the selected $\s{A}$ should have good design properties. Recall that $\s{A}$ is used for interpolation. The corresponding design properties include space-filling properties (Santner, Williams and Notz 2003) and low-dimensional projection properties (Joseph, Gul, and Ba 2015). Here we recommend selecting $\s{A}=\{\v{a}_1,\ldots,\v{a}_m\}$ among $m$-subsets of $\s{X}$ by minimizing the following criterion\begin{equation}\label{c}\mathrm{c}(\s{A})=\max_{1\leqslant i<j\leqslant m}\sum_{l=1}^d\frac{1}{|a_{il}-a_{jl}|},\quad\v{a}_i=(a_{i1},\ldots,a_{id})',\ i=1,\ldots,m,\end{equation}which is easy to compute and balances space-filling and low-dimensional projection properties relatively well (Mu and Xiong 2018). It should be noted that the optimal design property usually leads to feasible estimators instead of optimal estimators. In addition, our experience indicates that the performance of the proposed method does not heavily rely on the selection of $\s{A}$. In practice, we can randomly generate many $m$-subsets and select the one with the minimum value of the criterion \eqref{c}.

As mentioned above, we sometimes need to add some knots. A feasible sequential method is as follows.
Let $\hat{\vg{\gamma}}$ be the current estimator of $\vg{\gamma}$ based on current $\s{A}$.
We select the next point corresponding to the maximum residual, i.e., $\v{a}_{m+1}={\mathrm{arg}}\max_{\v{x}_i\in\s{X}\setminus\s{A}}\big[y_i-\mathcal{I}(\v{x}_i;\,\s{A},\hat{\vg{\gamma}})\big]^2$. With the updated $\s{A}$, we re-estimate $f$ and evaluate the estimator by a CV-like criterion. Repeat the above steps until the criterion becomes nondecreasing or stable.

This paper focuses on the situations where $\v{x}_1,\ldots,\v{x}_n$ are scattered on a hypercube. When the data are irregularly spaced, the above selection of $\s{A}$ should be modified based on the corresponding effective design methods for interpolation (Pratola et al. 2017). This issue is valuable to further investigate in the future.

\subsection{Estimation of kernel parameters}\label{subsec:kp}

Popular kernels involve several kernel parameters; see $\theta_j$'s in the Gaussian kernel \eqref{gk} and $(\nu,\phi)$ in the Mat\'{e}rn kernel \eqref{mk} for examples. The performance of Gaussian process interpolators, and thus GPRR estimators, depends on the selection of these parameters. We sometimes need to estimate them.

Let $\vg{\theta}$ denote the vector of unknown kernel parameters in a kernel/correlation function $R(\cdot\,|\,\vg{\theta})$. Accordingly, rewrite the Gaussian process interpolator \eqref{fity} as \begin{equation}\label{fitwth}\mathcal{I}_{\mathrm{GP}}(\v{x};\,\s{A},\vg{\gamma})=\vg{\gamma}'\v{b}(\v{x};\vg{\theta}),\end{equation}where the corresponding terms in $\v{b}(\v{x};\vg{\theta})$ depend on $\vg{\theta}$. A straightforward method is to estimate $\vg{\theta}$ along with the tuning parameter $\lambda$ by minimizing the GCV criterion \eqref{gcv0}. Here we provide another method for the cases where there is no penalty with $m\ll n$.
This method estimates $\vg{\theta}$ along with $\vg{\gamma}$ in \eqref{fitwth} by the method of least squares, i.e.,
\begin{equation}\label{se}\min_{\vg{\gamma},\vg{\theta}}\frac{1}{n}\sum_{i=1}^n\big[y_i-\vg{\gamma}'\v{b}\left(\v{x}_i;\vg{\theta}\right)\big]^2.\end{equation}
The block coordinate descent algorithm (Tseng 2001) can be used to solve \eqref{se}. In the $k$th iteration of this algorithm, $\vg{\theta}^{(k)}$ is obtained by minimizing the objective function for given $\vg{\gamma}=\vg{\gamma}^{(k-1)}$, and similarly $\vg{\gamma}^{(k)}$ is obtained by \eqref{krrf} for given $\vg{\theta}=\vg{\theta}^{(k)}$. Our experience shows that this method usually outperforms the GCV-based method.

\section{Numerical examples for GPRR}\label{sec:simu}

\subsection{Comparisons of methods with $\s{A}=\s{X}$}\label{subsec:fk}

This subsection considers the following test functions on $[0,1]^d$,	
	\begin{eqnarray*}&&\mathrm{(I)}\ f(\v{x})=\sum\limits_{j=1}^d jx_j^2,\\
		&&\mathrm{(II)}\ f(\v{x})=-20\exp \left(-\dfrac{1}{5}\sqrt{\dfrac{1}{d}\sum\limits_{j=1}^{d}x_j^2}\right)-\exp \left	 (\dfrac{1}{d}\sum\limits_{j=1}^d 2 \pi x_j\right)+20+\exp(1),
		\\&&\mathrm{(III)}\ f(\v{x})=-\left(\sum_{j=1}^dx_j\right)\exp\left(-\sum_{j=1}^dx_j^2\right).\end{eqnarray*}
	Model $\mathrm{(I)}$ is known as weighted sphere model, model $\mathrm{(II)}$ is Ackley's model, and model $\mathrm{(III)}$ is Yang's model (Yang 2010). Let $\v{x}_{1},\ldots,\v{x}_n$ be independently generated from the uniform distribution on $[0,1]^d$. The random errors $\varepsilon_1,\ldots,\varepsilon_n$ in \eqref{nr} are independently generated from $N(0,1)$. We consider the cases of $d=2,4$ and $n=200,500$, and use $N=10000$ test data from the uniform distribution on $[0,1]^d$ to compute the squared test error $$\frac{1}{N}\sum_{i=1}^N[\hat{f}(\v{x}_{\mathrm{test},i})-f(\v{x}_{\mathrm{test},i})]^2.$$The mean squared test error (MSE) is obtained over 100 repetitions. Three methods, KRR, GPR with linear regression terms, and GPRR, are compared. In GPRR, we use $\s{A}=\s{X}$ and the Gaussian process interpolator \eqref{fity} with $\v{g}(\v{x})=(1,x_1,\ldots,x_d)'$. In all the three methods, we use the Gaussian kernel \eqref{gk} with $\theta_1=\cdots=\theta_d=12.5$ (Li, Liu, and Zhu 2007), and use GCV to select the tuning parameters. The results are presented in Table \ref{tb:peu}.
It can be seen that GPRR is the best for all the cases.

\begin{table}[t]\centering\footnotesize
\footnotesize
\caption{Comparisons of MSEs in Section \ref{subsec:fk} (standard deviations in parentheses)}
\vspace{2mm}(I)\\\begin{tabular}{l c c c c c}
\hline\hline
   & \multicolumn{2}{c}{$d=2$} & & \multicolumn{2}{c}{$d=4$} \\\cline{2-3}\cline{5-6}
   & $n=200$ & $n=500$ & & $n=200$ & $n=500$  \\ \hline
  KRR  & 0.0901 (0.0328) & 0.0419 (0.0113)& & 1.3127 (0.1731) & 0.6162 (0.0637) \\
  GPR  & 0.0786 (0.0282) & 0.0391 (0.0117)& & 0.1834 (0.0279) & 0.1575 (0.0214) \\
  GPRR & 0.0373 (0.0202) & 0.0185 (0.0074)& & 0.1371 (0.0229) & 0.0888 (0.0146) \\
 \hline
\end{tabular}\\[2mm]
(II)\\\begin{tabular}{l c c c c c}
\hline\hline
   & \multicolumn{2}{c}{$d=2$} & & \multicolumn{2}{c}{$d=4$} \\\cline{2-3}\cline{5-6}
   & $n=200$ & $n=500$ & & $n=200$ & $n=500$  \\ \hline
  KRR  & 0.1258 (0.0420) & 0.0585 (0.0150)& & 0.9044 (0.1096) & 0.4346 (0.0467) \\
  GPR  & 0.0885 (0.0302) & 0.0452 (0.0118)& & 0.1780 (0.0286) & 0.1539 (0.0123) \\
  GPRR & 0.0810 (0.0288) & 0.0414 (0.0117)& & 0.1119 (0.0268) & 0.0741 (0.0103) \\
 \hline
\end{tabular}\\[2mm]
(III)\\\begin{tabular}{ l c c c c c}
\hline\hline
   & \multicolumn{2}{c}{$d=2$} & & \multicolumn{2}{c}{$d=4$} \\\cline{2-3}\cline{5-6}
   & $n=200$ & $n=500$ & & $n=200$ & $n=500$  \\ \hline
  KRR  & 0.0391 (0.0203) & 0.0193 (0.0066)& & 0.0991 (0.0219) & 0.0638 (0.0109) \\
  GPR  & 0.0746 (0.0260) & 0.0412 (0.0101)& & 0.1494 (0.0299) & 0.1389 (0.0184) \\
  GPRR & 0.0259 (0.0187) & 0.0120 (0.0055)& & 0.0370 (0.0193) & 0.0176 (0.0071) \\
 \hline
\end{tabular}\label{tb:peu}
 \end{table}

\subsection{Comparisons of methods with $m<n$}\label{subsec:pk}

\begin{table}
\caption{\label{tab:bh}Inputs of the borehole model} \centering\vspace{2mm}
\begin{tabular}{lll} \hline \quad \quad \quad \quad \quad \quad \quad \ Input&\quad\quad  Range& Unit
\\\hline$r_w$:\quad radius of borehole&[0.05, 0.15]&${\mathrm{m}}$
\\$r$:\quad radius of influence&[100, 50000]&${\mathrm{m}}$
\\$T_u$:\quad transmissivity of upper aquifer&[63070, 115600]&${\mathrm{m^2/yr}}$
\\$H_u$:\quad potentiometric head of upper aquifer&[990, 1110]&${\mathrm{m}}$
\\$T_l$:\quad transmissivity of lower aquifer&[63.1, 116]&${\mathrm{m^2/yr}}$
\\$H_l$:\quad potentiometric head of lower aquifer&[700, 820]&${\mathrm{m}}$
\\$L$:\quad length of borehole&[1120, 1680]&${\mathrm{m}}$
\\$K_w$:\quad hydraulic conductivity of borehole&[1500, 15000]&${\mathrm{m/yr}}$
\\\hline \end{tabular}
\end{table}

We now compare GPRR of $m<n$ and existing methods using the regression function \begin{eqnarray}\label{bh}y=\frac{2\pi T_u(H_u-H_l)}{\displaystyle \log(r/r_w)\Big[1+\frac{\displaystyle
2LT_u}{\displaystyle \log(r/r_w)r_w^2K_w}+T_u/T_l\Big]},\end{eqnarray}which is the borehole model (Worley 1987) and let the random errors $\varepsilon_1,\ldots,\varepsilon_n$ in \eqref{nr} be independently generated from $N(0,1)$. The borehole model describes the flow of water through a borehole drilled from the ground surface through two aquifers, and has been widely used in the literature for illustrating various methods; see Morris, Mitchell, and Ylvisaker (1993), Mease and Bingham (2006), and Xiong, Qian, and Wu (2013) among many others. Table \ref{tab:bh} presents the eight inputs of the model and their ranges and units.

In this simulation, the training data of $n=5000$ and $n=10000$ are generated by \eqref{bh} with inputs from the uniform distribution on their ranges. We use $N=20000$ test data from the same uniform distribution to compute the mean squared test errors. Three methods, the Nystr\"{o}m method (Williams and Seeger 2000), SPGP (Snelson and Ghahramani 2006), and GPRR, are compared. For the two sample sizes, consider $m=10d=80$ and $m=160$. We repeat 40 times. For each time, we randomly take an $m$-subset $\s{A}$ of $\s{X}$ for 50 times. In the Nystr\"{o}m method, the $m$-rank approximation, corresponding to $\s{A}$, of the $n\times n$ covariance matrix is used in GPR with linear regression terms. In SPGP, $\s{A}$ is taken as the set of pseudo inputs. In GPRR, let $\s{A}$ be the set of knots, and we use the Gaussian process interpolator \eqref{fity} with $\v{g}(\v{x})=(1,x_1,\ldots,x_d)'$ and the Gaussian kernel \eqref{gk} with the kernel parameters being given by \eqref{se}. For each $\s{A}$ in the 50 times, we compute the MSEs of the three methods and the corresponding standard deviations. Consequently the mean MSEs (MMSEs) and mean standard deviations (MSTDs) are computed over the 40 repetitions; see Table \ref{tb:pk}.

\begin{table}[t]\centering\footnotesize
\footnotesize
\caption{Comparisons of MMSEs and MSTDs in Section \ref{subsec:pk}}\vspace{2mm}
\begin{tabular}{l c c c c c}
\hline\hline
   & \multicolumn{2}{c}{$n=5000$} & & \multicolumn{2}{c}{$n=10000$} \\\cline{2-3}\cline{5-6}
   & $m=80$ & $m=160$ & & $m=80$ & $m=160$  \\ &MMSE (MSTD)&MMSE (MSTD)&&MMSE (MSTD)&MMSE (MSTD)\\\hline
  Nystr\"{o}m  & 42.683 (84.952) & 38.060 (82.590)& & 68.895 (79.955) & 25.955 (40.797) \\
  SPGP         & 10.211 (35.183) & 10.053 (17.518)& & 9.7605 (4.0514) & 8.9542 (3.1567) \\
  GPRR         & 1.2475 (0.8114) & 0.6749 (0.5756)& & 0.9020 (0.4101) & 0.6402 (0.2886) \\
 \hline
\end{tabular}\label{tb:pk}
 \end{table}

We can see from the results that, the Nystr\"{o}m method does not give good approximations for relatively small $m$, SPGP seems not good either, and the proposed GPRR method is much better than them. Furthermore, the small MSTDs of GPRR indicate that this method does not heavily depend on the selection of $\s{A}$.

\subsection{A real data example}\label{subsec:ccpp}

We apply our GPRR method to analyze a real dataset. The dataset contains 9568 data points collected from a Combined Cycle Power Plant (CCPP) over 6 years (2006-2011) (T\"{u}fekci 2014), when the power plant was set to work with full load. Features consist of hourly average ambient variables Temperature, Ambient Pressure, Relative Humidity, and Exhaust Vacuum to predict the net hourly electrical energy output of the plant. Among those 9568 data, the first 9000 of them are set as the training sample and the remaining 568 of them are set to be the test sample.

\begin{figure}[tbp]\begin{center}
\scalebox{0.6}[0.6]{\includegraphics{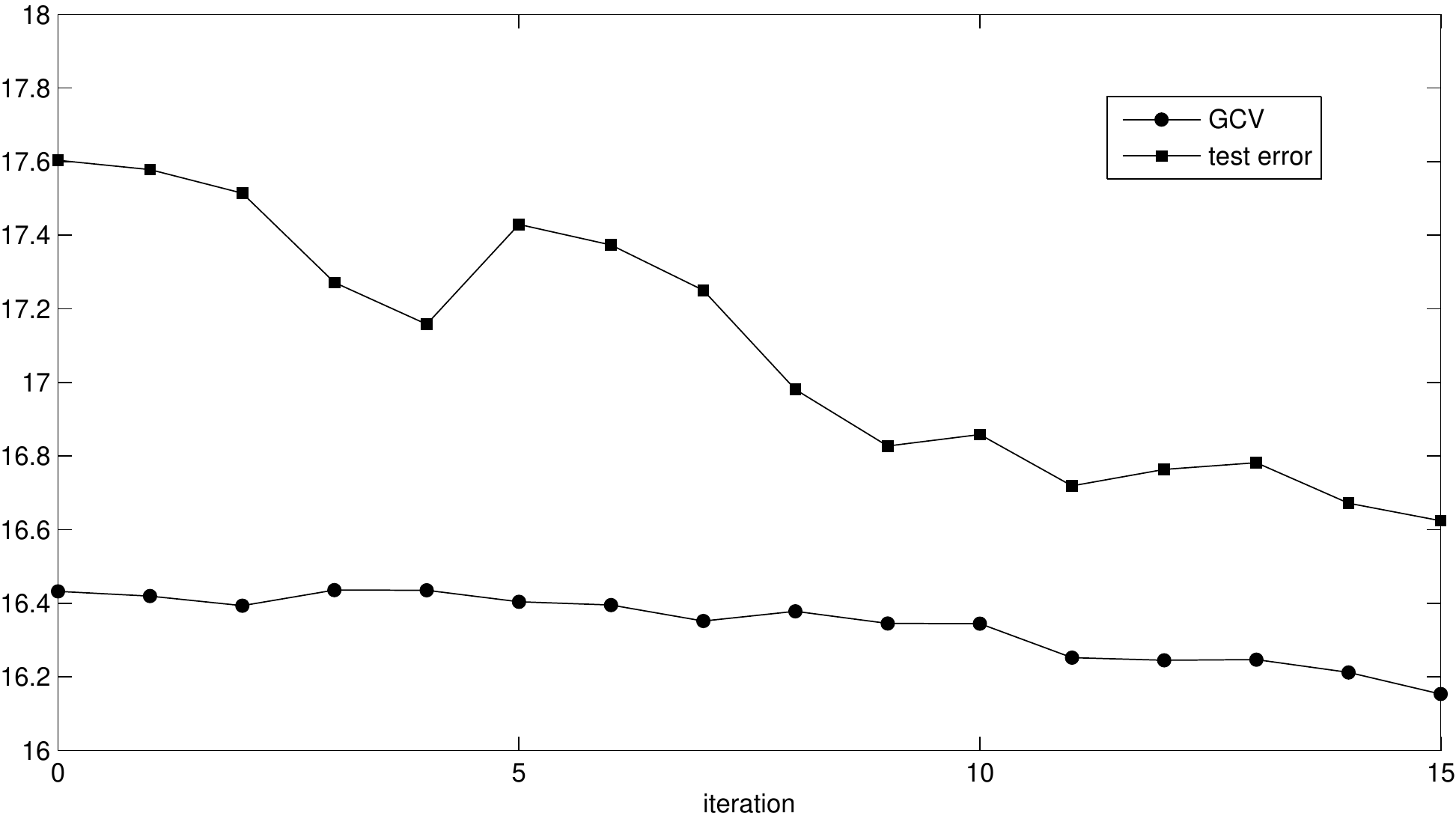}}
\end{center}
\caption{The sequential method in Section \ref{subsec:ccpp}.}\label{fig:seq}
\end{figure}

For $m=10d=40$, the knot set $\s{A}$ is selected by \eqref{c} among 20000 randomly generated subsets. With this $\s{A}$, we conduct the three methods in Section \ref{subsec:pk}, the Nystr\"{o}m method, SPGP, and GPRR, and the test errors of them are 86.6040, 30.5873, and 17.6035, respectively. This result still shows that the GPRR method with a small number of knots has satisfactory accuracy.

We then use the sequential procedure in Section \ref{subsec:alg} to improve on the GPRR estimator. The values of GCV and test errors of 15 iterations are reported in Figure \ref{fig:seq}. It can be seen that they consistently decrease on the whole.

\section{Discussion}\label{sec:dis}

In this paper we have proposed the reconstruction approach for nonparametric regression. Theoretical basis behind the proposed approach is that interpolators of sufficiently smooth functions usually yield negligible errors compared to statistical errors. We have shown its several features.
\begin{description}

\item[](a). This approach is easy to understand and to implement. The parameters in it have intuitive interpretation: they are function values at selected knots. This point is much different from other parameterization methods, and can bring some convenience in its implementation.

\item[](b). It provides an interpolation angle to examine existing methods. Several popular methods can be viewed as its special cases. In particular, for some methods with $n$ basis functions like GPR, it yields surrogates with much less basis functions. Therefore, the proposed approach is suitable for large scale problems.

\item[](c). It allows flexible implementation with different sizes of knot sets. Our numerical experiments show that the proposed approach has satisfactory performance for both small/moderate and large scale problems.

\item[](d). It systematically connects the two important areas: interpolation and regression. By the reconstruction idea, a regression problem can be handled based on an efficient interpolator. Other possible applications include classification, density estimation, quantile regression, and semiparametric regression. The reconstruction approach broadens the applications of interpolation in statistics and machine learning.

\end{description}

\section*{Acknowledgements}

I thank the Editors and referees for constructive comments which lead to a significant improvement of this paper. I also thank Lulu Kang for helpful discussion.
This work is supported by the National Natural Science Foundation of China (Grant No. 11671386, 11871033), and Key Laboratory of Systems and Control, CAS.

\vspace{1cm} \noindent{\Large\bf References}

{\begin{description}

\footnotesize

\item{}
Berlinet, A. and Thomas-Agnan, C. (2004). \textit{Reproducing Kernel Hilbert Spaces in Probability and Statistics}, Springer.

\item{}
Box, G. E. P. and Draper, N. R. (2007). \textit{Response Surfaces, Mixtures, and Ridge Analyses}, Second Edition, John Wiley \& Sons.


\item{}
Breiman, L., Friedman, J., Olshen, R. and Stone, C. (1984). \textit{Classification and Regression Trees}, Wadsworth.


\item{}
Celant, G. and Broniatowski, M. (2016). \textit{Interpolation and Extrapolation Optimal Designs V1: Polynomial Regression and Approximation Theory}, ISTE Ltd. 

\item{}
Cressie, N. A. C. (2015). \textit{Statistics for Spatial Data}, Revised Edition. John Wiley \& Sons.

\item{}
Cressie, N. A. C. and Johannesson, G. (2008). Fixed rank kriging for very large spatial data sets. \textit{Journal of The Royal Statistical Society Series B-statistical Methodology}, {\bf 70}, 209--226.

\item{}
Dasarathy, B. (1991). \textit{Nearest neighbor pattern classification techniques}, IEEE Computer Society Press.


\item{}
De Boor, C. (1978). \textit{A Practical Guide to Splines}, Springer-Verlag.


\item{}
Eilers, P. H. C. and Marx, B. D. (1996). Flexible smoothing with $B$-splines and penalties, \textit{Statistical Science}, {\bf 11}, 89--102.

\item{}
Eubank, R. L. (1999). \textit{Nonparametric Regression and Spline Smoothing}, Second Edition, Marcel Dekker.  

\item{}
Fan, J. and Gijbels, I. (1996). \textit{Local Polynomial Modelling and Its Applications}, Chapman \& Hall/CRC.



\item{}Friedman, J. H. (2001). Greedy function approximation: A gradient boosting machine, \textit{Annals of Statistics}, {\bf 29}, 1189--1232.

\item{}
Friedman, J. H., Hastie, T., and Tibshirani, R. (2008). \textit{The Elements of Statistical Learning}, Second Edition, Springer.

\item{}
Furrer, R., Genton, M., and Nychka, D. (2006). Covariance tapering for interpolation of large spatial datasets, \textit{Journal of Computational and Graphical Statistics}, {\bf 15}, 502--523.

\item{}
Golub, G. H., Heath, M., and Wahba, G. (1979). Generalized cross-validation as a method for choosing a good ridge parameter, \textit{Technometrics}, {\bf 21}, 215--223.

\item{}
Goodfellow, I., Bengio, Y., and Courville, A. (2016). \textit{Deep Learning}, MIT Press.

\item{}
Gramacy, R. B. and Apley, D. W. (2015). Local Gaussian process approximation for large computer experiments, \textit{Journal of Computational and Graphical Statistics}, {\bf 24}, 561--578.

\item{}
Gu, C. (2002). \textit{Smoothing Spline ANOVA Models}, Springer.


\item{}
Hall, P. and Turlach, B. A. (1997). Interpolation methods for adapting to sparse design in nonparametric regression, \textit{Journal of the American Statistical Association},
{\bf 92}, 466--472.




\item{}
Joseph, V. R., Gul, E., and Ba, S. (2015). Maximum projection designs for computer experiments, \textit{Biometrika}, {\bf 102}, 371--380.

\item{}
Kang, L. and Joseph, V. R. (2016). Kernel approximation: from regression to interpolation, \textit{SIAM/ASA Journal on Uncertainty Quantification}, {\bf 4}, 112--129.

\item{}
Kung, S. Y. (2014). \textit{Kernel Methods and Machine Learning}, Cambridge University Press.


\item{}
Li, Y., Liu, Y., and Zhu, J. (2007). Quantile regression in reproducing kernel Hilbert space, \textit{Journal of the American Statistical Association},
{\bf 102}, 255--268.

\item{}
Loeppky, J. L, Sacks, J., and Welch, W. J. (2009). Choosing the sample size of a computer experiment: A practical guide. \textit{Technometrics}, {\bf 51}, 366--376.


\item{}
Mease, D. and Bingham, D. (2006). Latin hypercube sampling for computer experiments, \textit{Technometrics}, {\bf 48}, 467--477.

\item{}
Mu, W. and Xiong, S. (2018). A class of space-filling designs and their projection properties, \textit{Statistics and Probability Letters}, In Press.

\item{}
Pratola, M. T., Harari, O., Bingham, D., and Flowers, G. E. (2017). Design and analysis of experiments on nonconvex regions, \textit{Technometrics},  {\bf 59}, 36--47.

\item{}
Ramsay, J.O., Hooker, G., and Graves, S. (2009). \textit{Functional Data Analysis with R and MATLAB}, Springer.

\item{}
Rasmussen, C. E. and Williams, C. K. I. (2006). \textit{Gaussian Processes for Machine Learning}, MIT Press.


\item{}
Sang, H. and Huang, J. Z. (2012). A full scale approximation of covariance functions for large spatial data sets. \textit{Journal of The Royal Statistical Society Series B-statistical Methodology}, {\bf 74}, 111--132.

\item{}
Santner, T. J., Williams, B. J., and Notz, W. I. (2003). \textit{The Design and Analysis of Computer Experiments}, Springer. 

\item{}
Saunders, C., Gammerman, A., and Vovk, V. (1998). Ridge regression learning algorithm in dual variables, \textit{ICML}, {\bf 98}, 515--521.


\item{}
Snelson, E. and Ghahramani, Z. (2006). Sparse Gaussian processes using pseudo-inputs. In \textit{Advances in Neural Information Processing Systems}, 1257--1264.

\item{}
Stewart, G. W. (1996). \textit{Afternotes on Numerical Analysis}, Society for Industrial and Applied Mathematics.   

\item{}
Stewart, G. W. (1998). \textit{Afternotes goes to Graduate School: Lectures on Advanced Numerical Analysis}, Society for Industrial and Applied Mathematics.   

\item{}
Tibshirani, R. J. (2014). Adaptive piecewise polynomial estimation via trend filtering, \textit{Annals of Statistics}, {\bf 42}, 285--323.

\item{}
Toraichi, K., Katagishi, K., Sekita, I., and Mori, R. (1987). Computational complexity of spline interpolation, \textit{International Journal of Systems Science}, {\bf 18}, 945--954.

\item{}
Tseng, P. (2001). Convergence of a block coordinate descent method for nondifferentiable minimization, \textit{Journal of Optimization Theory and Applications}, {\bf 109}, 475--494.

\item{}
T\"{u}fekci, P. (2014). Prediction of full load electrical power output of a base load operated combined cycle power plant using machine learning methods, \textit{International Journal of Electrical Power \& Energy Systems}, {\bf 60}, 126--140.



\item{}
van Wieringen, W. N. (2018). Lecture notes on ridge regression, \textit{arXiv preprint arXiv:1509.09169}.


\item{}
Wendland, H. (2004). \textit{Scattered Data Approximation}, Cambridge University Press.

\item{}
Williams, C. K. and Seeger, M. (2000). Using the Nystr\"{o}m method to speed up kernel machines. In \textit{Advances in Neural Information Processing Systems}, 682--688.

\item{}
Worley, B. A. (1987). Deterministic uncertainty analysis, \textit{Technical Report ORNL-6428, Oak Ridge National Laboratories}.

\item{}
Xie, M. Y. and Fang, K. T. (2000). Admissibility and minimaxity of the uniform design measure in nonparametric regression model, \textit{Journal of Statistical Planning and Inference}, {\bf 83}, 101--111.

\item{}
Xiong, S., Qian, P. Z. G., and Wu, C. F. J. (2013). Sequential design and analysis of high-accuracy and low-accuracy computer codes, \textit{Technometrics}, {\bf 55}, 37--46.

\item{}
Yang, X.-S. (2010). Test Problems in Optimization, \textit{Engineering optimization}, ed. X.-S. Yang, John Wiley and Sons, Inc.


\end{description}}

\end{document}